\documentclass[11pt,twoside]{article} 
\usepackage{acta-info}
\usepackage{euler}
\usepackage{graphicx}
\newcommand{\vol}{\textbf{2,} 1 (2010) 40--50}   
\newcommand{\amss}{\amssubj{%
68N99
}}     
\newcommand{\CRs}{\CRsubj{%
D.2.8 
}}   
\newcommand{\keyw}{\keywords{%
program comprehension, domain knowledge, program quality, software measurement
}}

\begin{document}

\title{%
Automatic derivation of domain terms and concept location based on the analysis of the identifiers
}
\maketitle

\oneauthor{%
\href{http://hornad.fei.tuke.sk/kpi/person/vaclavik/dcicard.php}{Peter V\'aclav\'{\i}k} 
}{%
\href{http://www.tuke.sk}{Technical University of Ko\v{s}ice}\\
\href{http://www.fei.tuke.sk/}{Faculty of Electrical Engineering and Informatics}\\
\href{http://kpi.fei.tuke.sk/}{Department of Computers and Informatics}
}{%
 \href{mailto:Peter.Vaclavik@tuke.sk}{Peter.Vaclavik@tuke.sk}
}


\twoauthors{%
\href{http://hornad.fei.tuke.sk/kpi/person/poruban/dcicard.php}{Jaroslav Porub\"an} 
}{%
\href{http://www.tuke.sk}{Technical University of Ko\v{s}ice}\\
\href{http://www.fei.tuke.sk/}{Faculty of Electrical Engineering and Informatics}\\
\href{http://kpi.fei.tuke.sk/}{Department of Computers and Informatics}
}{%
 \href{mailto:Jaroslav.Poruban@tuke.sk}{Jaroslav.Poruban@tuke.sk}
}{%
Marek Mezei
}{%
\href{http://www.tuke.sk}{Technical University of Ko\v{s}ice}\\
\href{http://www.fei.tuke.sk/}{Faculty of Electrical Engineering and Informatics}\\
\href{http://kpi.fei.tuke.sk/}{Department of Computers and Informatics}
}{%
 \href{mailto:marekmezei@gmail.com}{marekmezei@gmail.com}
}


\short{%
P. V\'aclav\'{\i}k, J. Porub\"an, M. Mezei
}{%
Domain terms and concept location based on identifiers' analysis
}

\begin{abstract}
Developers express the meaning of the domain ideas in specifically selected identifiers and comments that form the target implemented code. Software maintenance requires knowledge and understanding of the encoded ideas. This paper presents a way how to create automatically domain vocabulary. Knowledge of domain vocabulary supports the comprehension of a specific domain for later code maintenance or evolution. We present experiments conducted in two selected domains: application servers and web frameworks. Knowledge of domain terms enables easy localization of chunks of code that belong to a certain term. We consider these chunks of code as ``concepts'' and their placement in the code as ``concept location''. Application developers may also benefit from the obtained domain terms. These terms are parts of speech that characterize a certain concept. Concepts are encoded in ``classes'' (OO paradigm) and the obtained vocabulary of terms supports the selection and the comprehension of the class' appropriate identifiers. We measured the following software products with our tool: JBoss, JOnAS, GlassFish, Tapestry, Google Web Toolkit and Echo2.
\end{abstract}


\section{Introduction}
Program comprehension is an essential part of software evolution and software maintenance: software that is not comprehended cannot be changed \cite{Rajlich,Samuelis,Samuelis-Szabo,Szabo-Samuelis}.

Among the earliest results are the two classic theories of program comprehension, called \textit{top-down} and \textit{bottom-up} theories \cite{Storey}. Bottom-up theory: Consider that understanding a program is obtained from source code reading and then mentally chunking or grouping the statements or control structures into higher abstract level, i.e. from bottom up. Such information is further aggregated until high-level abstraction of the program is obtained. Chunks are described as code fragments in programs. Available literature shows chunks to be used during the bottom-up approach of software comprehension. Chunks vary in size. Several chunks can be combined into larger chunks \cite{Aschwanden}. On the other hand, the top-down approach starts the comprehension process with a hypothesis concerning a high-level abstraction, which then will be further refined, leading to a hierarchical comprehension structure. The understanding of the program is developed from the confirmation or refutation of hypotheses.

An important task in program comprehension is to understand where and how the relevant concepts are located in the code. Concept location is the starting point for the desired program change. Concept location means a process where we assume that programmer understands the concept of the program domain, but does not know where is it located within the code. All domain concepts should map onto one or more fragments of the code. In other words, process of concept location is the process that finds that code-fragment \cite{Rajlich}.

Developers who are new to a project know little about the identifiers or comments in the source code, but it is likely that they have some knowledge about the problem domain of the software. In this paper, we  present a new way of  program comprehension that is based on naming of identifiers. When trying to understand the source code of a software system, developers usually start by locating familiar concepts in the source code. Keyword search is one of the most popular methods for this kind of task, but the success is strictly tied to the quality of the user queries and the words used to construct the identifiers and comments.

We present a way how to create a domain vocabulary automatically as a result of source code analysis. We classify the parts of speech and measure their occurrence in the source code.

\section{Motivation}
Domain level knowledge is important when programmers attempt to understand a program. Programmer inspects source code structure that is directed by identifiers. The quality and the ``orthogonality''  of the identifiers in the source code affects the time of program comprehension. Next kinds of quality could be measured:

\begin{enumerate}\addtolength{\itemsep}{-0.6\baselineskip}
	\item percentage of \textit{fullword} identifiers,
	\item percentage of \textit{abbreviations} and \textit{unrecognized} identifiers,
	\item percentage of \textit{domain terms} identified in the application.
\end{enumerate}

Percentage of full word identifiers is very important in the case of absence of documentation. The first two qualities could be derived directly from the source code toward common vocabulary.  We don't need any additional domain data source to get relevant results. The third quality is not derived directly. We need to make measurements in order to obtain domain vocabulary.

Usually we don't have domain terms of the analysed software product. The question is: \textit{How can we create  the vocabulary of terms for a particular domain?} In this paper we propose a way to derive it automatically.

Nowadays, the companies are affected by employee fluctuation, especially in the IT sector. Each company has ongoing projects in the phase of developing or maintenance. New developer participating  in the project has to understand project to solve the assigned task. Domain terms are usually in the specification. The transition from specification to implementation is bound usually to the transformation of terms. For example, if the specification contains word \textbf{car}, that word  could be changed to word \textbf{vehicle} in the implementation phase. In spite of the fact that the word \textbf{vehicle} is a hypernym of the word \textbf{car},  we cannot find the word ``vehicle'' by brute force through searching by keywords. That is the reason for looking for some statistical evidence that car is a vehicle. It means that there exists ``gap'' in the meanings between the words used in specification and implementation. Our goal is to eliminate partly ``this kind'' of gap.

Developers of new software products may put another question: \textit{What kind of parts of speech is usually used for a particular category of identifiers?} We can measure it directly from the source code. We can also find, if the rules are domain specific or generally applicable.

\section{Methodology of programm inspection}

Full word identifiers provide better comprehension then single letters or abbreviations \cite{Lawrie}. It is the reason why we want to provide a tool for measurement of this aspect of program quality. We use the WordNet database of words to identify the potential domain terms.

\begin{figure}[!ht]
\centering
\includegraphics*[scale=0.6]{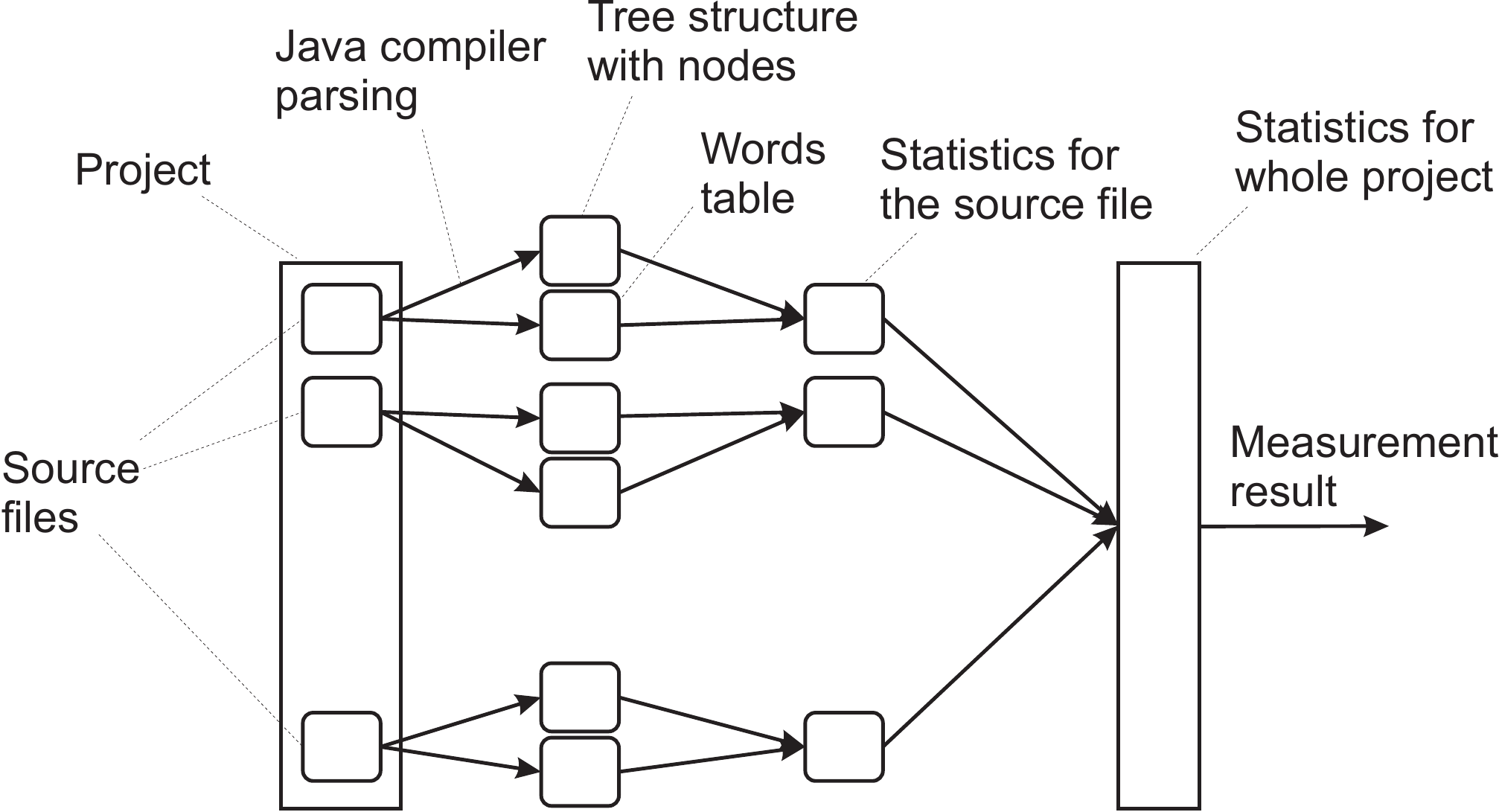}
\caption{The methodology of program inspection}
\label{fig:architecture}
\end{figure}

We apply our tool to well-known open-source projects. They belong to two domains: domain of application servers and domain of web frameworks. Each project consists of a set of source code files. We examine every source file separately. Based on information we have got by source files analysis we make measurements for the whole project.
Our measurements follow the scheme shown on the Fig. \ref{fig:architecture}.
\begin{enumerate}\addtolength{\itemsep}{-0.6\baselineskip}
	\item First, we parse every source file using Java compiler. We build a tree structure of nodes. Each node belongs to one of the next types: \textit{class}, \textit{method}, \textit{method parameter} or \textit{class variable}. Then we process the names of each identified node. Name processing consists of splitting the name according to common naming conventions. For example, ``setValue'' is split into ``set'' and ``value" words. After then we put all identified words into a table.
	\item As a second step, we produce statistics for the source file. We examine which word belongs to class variable, method parameter, method or class and also we try to assign part of speech to the words.
\end{enumerate}

After source files analysis mentioned in previous steps we produce statistics for the whole project: we build a set of words containing all words used in the source files, and also we build a set of words used in the variables (class variables and method parameters), methods and classes. The set of words used in project will represent the software vocabulary for the particular project.

The software domain vocabulary represents the intersection of all software vocabularies of all software products of the same domain. Not all identified words are suitable candidates for the inclusion into software domain vocabulary. It is expected to apply filters in a process of source code analysis. So, the reason behind filtering is to eliminate terms that are irrelevant regarding the domain. As a final result we obtain a set of words ordered by occurrence. We obtain domain vocabulary as well as potential domain vocabulary (words are not identified in all measured software products).

As was mentioned in the previous section, the categorization in accordance to the parts of speech is expected in the experiment. It induces another problem: one word can belong to more parts of speech (e.g. ``good'' is adjective as well as noun). WordNet provides help in disambiguation and classification of words.

\section{Experiments based on word analysis}
WordNet provides a database of the most used words in the parts of speech. As was mentioned earlier, we have developed a tool to measure results in the graph, tabular and textual form. The tool's input is the project's source code. To present it we decide to inspect software products of two application domains:
\begin{itemize}\addtolength{\itemsep}{-0.6\baselineskip}
	\item Java EE application server,
	\item Web framework.
\end{itemize}

We have selected next Java EE application servers:
\begin{itemize}\addtolength{\itemsep}{-0.6\baselineskip}
	\item JOnAS 4.10.3 (\href{http://jonas.ow2.org/}{http://jonas.ow2.org/}),
	\item JBoss  5.0.1.GA (\href{http://www.jboss.org/jbossas}{http://www.jboss.org/jbossas}),
	\item GlassFish Server v2.1 (\href{https://glassfish.dev.java.net/}{https://glassfish.dev.java.net/}).
\end{itemize}

and web frameworks:
\begin{itemize}\addtolength{\itemsep}{-0.6\baselineskip}
	\item Google Web Toolkit 1.5.3  (GWT)    (\href{http://code.google.com/intl/sk/webtoolkit/}{http://code.google.com/intl/sk/webtoolkit/}),
	\item Echo2 v2.1 (\href{http://echo.nextapp.com/site/echo2}{http://echo.nextapp.com/site/echo2}),
	\item Tapestry 5.0.18 (\href{http://tapestry.apache.org/}{http://tapestry.apache.org/}).
\end{itemize}

\begin{table}[!ht]
\begin{center}
\begin{tabular}{ | l | l | l | l || l | l | l | }
\hline			
                      & {\bf Glass.} & {\bf JOnAS} & {\bf JBoss} & {\bf Echo2} & {\bf GWT} & {\bf Tap.} \\
\hline                           
    Number of source  & 10553        & 3611        & 6448        & 402         & 593       & 1707       \\
    files & & & & & & \\
\hline                           
    Number of words   & 10229        & 4502        & 5055        & 1044        & 2584      & 2154       \\
\hline                           
    Number of         & 4297         & 2140        & 2932        & 903         & 1582      & 1714       \\
    recognized words  & (42\%)       & (48\%)      & (58\%)      & (86\%)      & (61\%)    & (80\%)     \\
\hline                           
    Number of not     & 5932         & 2362        & 2123        & 141         & 1002      & 440        \\
    recognized words  & (58\%)       & (52\%)      & (42\%)      & (14\%)      & (39\%)    & (20\%)     \\
\hline                           
    Number of nouns   & 2361         & 1311        & 1687        & 537         & 839       & 913        \\
                      & (55\%)       & (61\%)      & (58\%)      & (60\%)      & (53\%)    & (54\%)     \\
\hline                           
    Number of verbs   & 1259         & 542         & 842         & 229         & 484       & 526        \\
                      & (29\%)       & (25\%)      & (29\%)      & (25\%)      & (31\%)    & (30\%)     \\
\hline                           
    Number of         & 549          & 235         & 330         & 117         & 213       & 226        \\
    adjectives        & (13\%)       & (11\%)      & (11\%)      & (13\%)      & (13\%)    & (13\%)     \\
\hline                           
    Number of         & 128          & 52          & 73          & 20          & 46        & 49         \\
    adverbs           & (3\%)        & (3\%)       & (2\%)       & (2\%)       & (3\%)     & (3\%)      \\
\hline 
\end{tabular}
\caption{The number of recognized domain-terms for application servers and web frameworks} \label{tab:result}
\end{center}
\end{table}

Table \ref{tab:result} summarizes data for the selected products. Other kind of results obtained from our tool in tabular form gives us information about identified words that are parts of software vocabulary. Now, each domain has three sets of software vocabulary. In our measurement we have selected the 50 most used words of each software vocabulary for further analysis. Their intersection is a set of terms belonging to the domain vocabulary. Table \ref{tab:dict} presents the most used words and their occurrence. Words identified as domain terms are emphasized with bold letters. Potential domain terms recognized in two software products are emphasized with italic. Words that belong to only one software vocabulary are typed ordinary. Thanks to WordNet we can also identify semantically similar (synonyms, homonyms, hypernyms, and so on) words as a domain or potential domain term.	

\begin{table}[!ht]
\begin{center}
\begin{tabular}{ | l | l | l || l | l | l | }
\hline			
   {\bf Glass.} & {\bf JBoss}  & {\bf JOnAS}  & {\bf GWT}    & {\bf Echo2}    & {\bf Tap.}\\
\hline                           
   {\bf name}   & {\bf name}   & {\bf name}   & {\it type}   & action         & {\bf name} \\
   (28727)      & (11774)      & (6950)       & (1774)       & (801)          & (1653)\\
\hline                           
   {\bf value}  & {\bf test}   & {\bf ejb}    & {\bf name}   & {\it property} & {\bf value} \\
   (9067)       & (7332)       & (2570)       & (1168)       & (506)          & (1140)\\
\hline                           
   {\bf type}   & {\bf id}     & {\bf test}   & {\it method} & {\bf value}    & {\it type} \\
   (7550)       & (3169)       & (2179)       & (540)        & (475)          & (970)\\
\hline                           
   {\bf class}  & {\bf bean}   & {\bf id}     & {\bf class}  & {\it test}     & {\it class} \\
   (6953)       & (3154)       & (1602)       & (466)        & (395)          & (936)\\
\hline                           
   {\it object} & {\bf ejb}    & {\it home}   & {\it logger} & {\it component}& field \\
   (5277)       & (2962)       & (1437)       & (353)        & (369)          & (798)\\
\hline                           
   {\bf id}     & {\bf value}  & {\bf server} & {\bf value}  & {\it element}  & page \\
   (4266)       & (2885)       & (1379)       & (344)        & (347)          & (778)\\
\hline                           
   {\bf key}    & {\bf type}   & {\bf type}   & info         & {\bf id}       & {\it component} \\
   (4641)       & (2473)       & (1251)       & (254)        & (331)          & (776)\\
\hline                           
\end{tabular}
\caption{Application server and web framework domain terms recognition} \label{tab:dict}
\end{center}
\end{table}

\section{An experiment on concept location}

Within the next step we locate concepts encoded in keywords of the product. We use again WordNet for searching keywords. Programmers knows only the domain the software product it belongs to. They do not need to use exact words used in source code.

We present here an example of concept location. Lets suppose that somebody wants to change the algorithm for determining the parts of speech in our program. S/he needs to locate the concept of determining the parts of speech in the source code of the examined program. It is known that programmers and maintainers use different words to describe essentially the same or similar concepts \cite{Rajlich}. Therefore the use of full-text search for concept location is very limited. We will try to find concepts based on semantic search.

In our example we assume that a concept is the identifier of a method or a class. We want to find a fragment in the source code where the parts of speech are located. We will try to find this code fragment based on this key-phrase: ``\textit{find word form}''. For every keyword in our key-phrase we will make a database of related words -- words that are in some semantic relationship to the keyword. Then, we will try to locate code fragment in our source code, where at least 1 occurrence for every keyword is found. This process is shown on the Fig. \ref{fig:location}. However we are not looking only for the keywords itself, but also for semantically related words. In our example, as a result we find a method with this definition:

\begin{verbatim}
  public String getType(String word) {
    //Method source code
  }
\end{verbatim}

\begin{figure}[!ht]
\centering
\includegraphics*[scale=0.6]{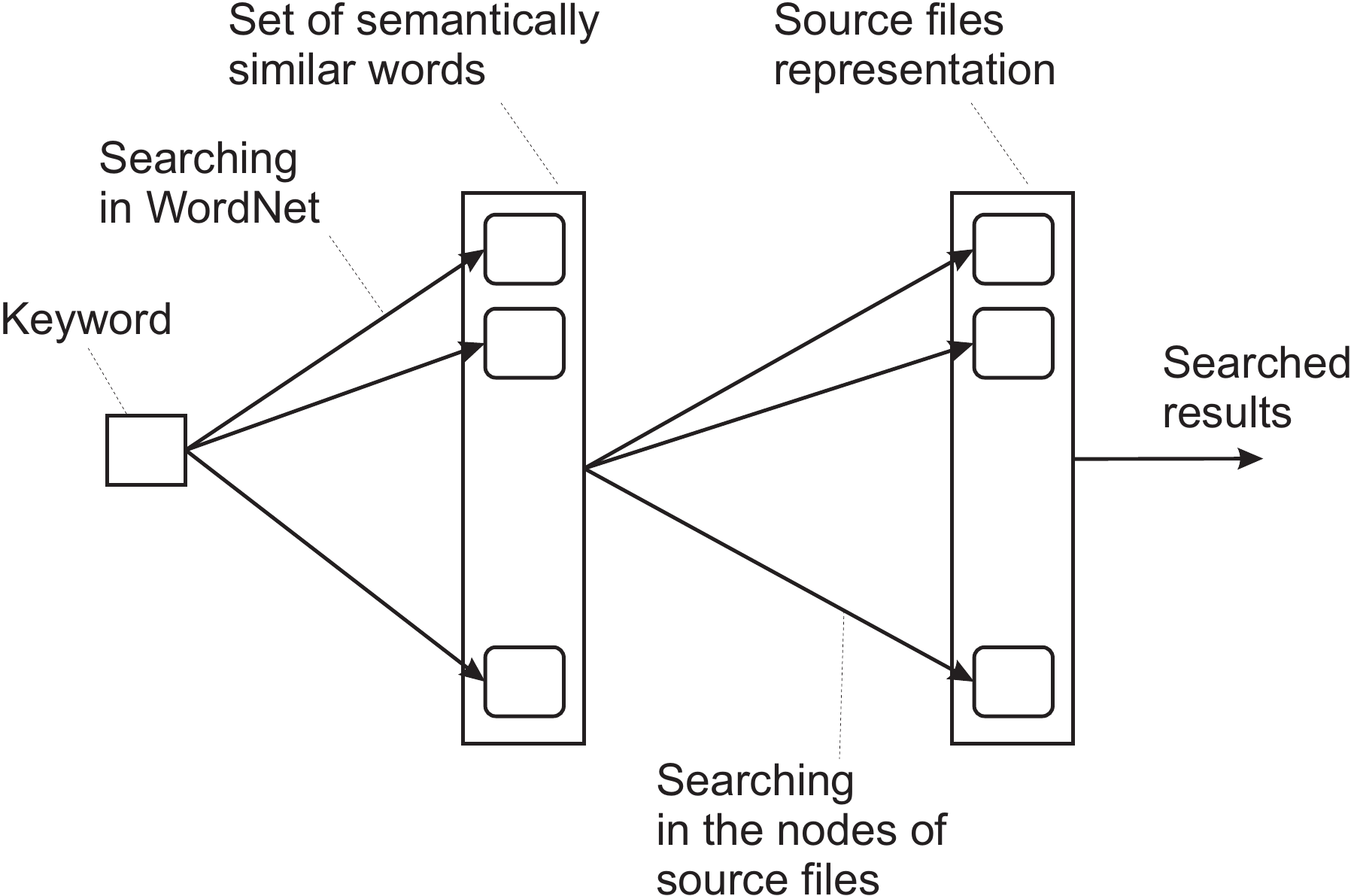}
\caption{Process of concept location}
\label{fig:location}
\end{figure}

We found the three keywords in this method definition based on these semantic relations:
\begin{enumerate}\addtolength{\itemsep}{-0.6\baselineskip}
	\item \textit{find--get}: ``get'' is a hypernym of ``find''. We found the word ``get'' in the method name.
	\item \textit{word}: we found the term ``word'' itself in the parameter name.
	\item \textit{form--type}: ``type'' is a hyponym of ``form''. We found the word ``type'' in the method name.
\end{enumerate}

We can see on this example that we could not locate this concept easily using fulltext search, but we can locate it using search based on semantic relations.

\section{Related and further steps}

\textbf{The study of software vocabularies.}  This study is focused on three research questions: (1) to what degree relate terms found in the source code to a particular domain?; (2) which is the preponderant source of domain terms: identifiers or comments?; and (3) to what degree are domain terms shared between several systems from the same domain? Within the studied software, we found that in average: 42\% of the domain terms were used in the source code; 23\% of the domain terms used in the source code are present in comments only, whereas only 11\% in the identifiers alone, and there is a 63\% agreement in the use of domain terms between any two software systems. They manually selected the most common concepts, based on several books and online sources. They chose 135 domain concepts. From the same resources, for each of these concepts one or more terms and standard abbreviations that describe the concept were manually selected and included in the domain vocabulary \cite{Haiduc}.

\textit{Our aim was to define the domain vocabulary automatically. Results from the experiments will be used to build domain vocabularies for other domains too. These vocabularies support more detailed automatic classification of software products. Our next experiments will include inspection of comments in the source code. This stream of research is strongly promoted by} \cite{Haiduc,Vinz1,Vinz2}.

\textbf{Concept location.}  One of the experiments in the area of mapping between source code and conceptualizations shared as ontology has been published in \cite{Ratiu}. The programs regard themselves as knowledge bases built on the programs' identifiers and their relations implied by the programming language. This approach extracts concepts from code by mapping the identifiers and the relations between them to ontology. As a result, they explicitly link the sources with the semantics contained in ontology. This approach is demonstrated using on the one hand the relations within Java programs generated by the type and the module systems and on the other hand the WordNet ontology.

\textit{We are locating concepts by keywords specified by programmers. Concept location is based on searching names in the identifiers that are in some relation to the specified keywords. This approach supports easier understanding of higher-level abstractions within the inspected application. We will work further on the concept visualization as well as on concept location refinement issues.}

\section{Conclusions}

We can conclude the experiment results briefly as follows:
\begin{itemize}\addtolength{\itemsep}{-0.6\baselineskip}
	\item In general, the most used parts of speech for all inspected element types are nouns (57\%).
	\item Application servers as well as GWT use a lot of not recognized words due to different identifiers.
	\item The most number of recognized words is used in Tapestry (80\%) and Echo2 (86\%) web frameworks. We can assume that the source code of both products could be well understandable.
	\item From the comprehension point of view the application servers are more complex than web frameworks.
	\item In spite of application servers' complexity, they are using more common domain terms. Application server domain vocabulary consists of other well-known terms like: ``context'', ``session'', ``service'', and so on.
	\item Concept location gives us opportunity to find source code fragments more efficiently and with better results than using classical keyword search.
\end{itemize}

\section*{Acknowledgement}
This work was supported by VEGA Grant No. 1/4073/07 - Aspect-oriented
Evolution of Complex Software System and by APVV Grant No. SK-CZ-0095-07 -
Cooperation in Design and Implementation of Language Systems. 



\bigskip
\rightline{\emph{Received: August 30, 2009  {\tiny \raisebox{2pt}{$\bullet$\!}} Revised February 25, 2010} }
\end{document}